\documentclass{article}

\usepackage[table,xcdraw]{xcolor}
\usepackage{amsmath}
\usepackage{wrapfig}
\usepackage[numbers, sort]{natbib}
\usepackage[utf8]{inputenc} 
\usepackage[T1]{fontenc}    
\usepackage{hyperref}       
\usepackage{url}            
\usepackage{booktabs}       
\usepackage{amsfonts}       
\usepackage{nicefrac}       
\usepackage{microtype}      
\usepackage{xcolor}         
\usepackage{multirow}
\usepackage{multicol}
\usepackage{graphicx} 
\usepackage{makecell}
\usepackage{utfsym}
\usepackage{tabularx}
\usepackage{footnote}
\usepackage{caption}
\usepackage{array}
\usepackage{authblk}
\usepackage{ragged2e}

\title{FUSU: A Multi-temporal-source Land Use Change Segmentation Dataset for Fine-grained Urban Semantic Understanding}

\author[1]{Shuai Yuan}
\author[2]{Guancong Lin}
\author[3]{Lixian Zhang}
\author[4]{Runmin Dong}
\author[4]{Jinxiao Zhang}
\author[1]{Shuang Chen}
\author[2]{Juepeng Zheng}
\author[5]{Jie Wang}
\author[6]{Haohuan Fu}
\affil[1]{The University of Hong Kong}
\affil[2]{Sun Yat-Sen University, Zhuhai}
\affil[3]{National Supercomputing Center, Shenzhen}
\affil[4]{Tsinghua University}
\affil[5]{Peng Cheng Labratory}
\affil[6]{Tsinghua Shenzhen International Graduate School, Tsinghua University}
\begin{document}

\maketitle

\begin{abstract}
Fine urban change segmentation using multi-temporal remote sensing images is essential for understanding human-environment interactions in urban areas. Although there have been advances in high-quality land cover datasets that reveal the physical features of urban landscapes, the lack of fine-grained land use datasets hinders a deeper understanding of how human activities are distributed across the landscape and the impact of these activities on the environment, thus constraining proper technique development. To address this, we introduce FUSU, the first fine-grained land use change segmentation dataset for Fine-grained Urban Semantic Understanding. FUSU features the most detailed land use classification system to date, with 17 classes and 30 billion pixels of annotations. It includes bi-temporal high-resolution satellite images with 0.2-0.5 $m$ ground sample distance and monthly optical and radar satellite time series, covering 847 $km^2$ across five urban areas in the southern and northern of China with different geographical features. The fine-grained land use pixel-wise annotations and high spatial-temporal resolution data provide a robust foundation for developing proper deep learning models to provide contextual insights on human activities and urbanization. To fully leverage FUSU, we propose a unified time-series architecture for both change detection and segmentation. We benchmark FUSU on various methods for several tasks. Dataset and code are available at: \textit{https://github.com/yuanshuai0914/FUSU}.
\end{abstract}

\section{Introduction}

Urban areas, housing 57\% of the world's population on just 3\% of global land, are dynamic hubs of human activity \citep{seto2012global}. The scale and rapid pace of current urbanization, encompassing both internal dynamics and population growth, position urban areas as a crucial catalyst of global climate change and vice versa \cite{seto2012urban}. Therefore, proper observation and monitoring of urban changes are crucial for modeling human-nature interactions.

\begin{figure*}[t]
    \centering
    \includegraphics[width=0.95\linewidth]{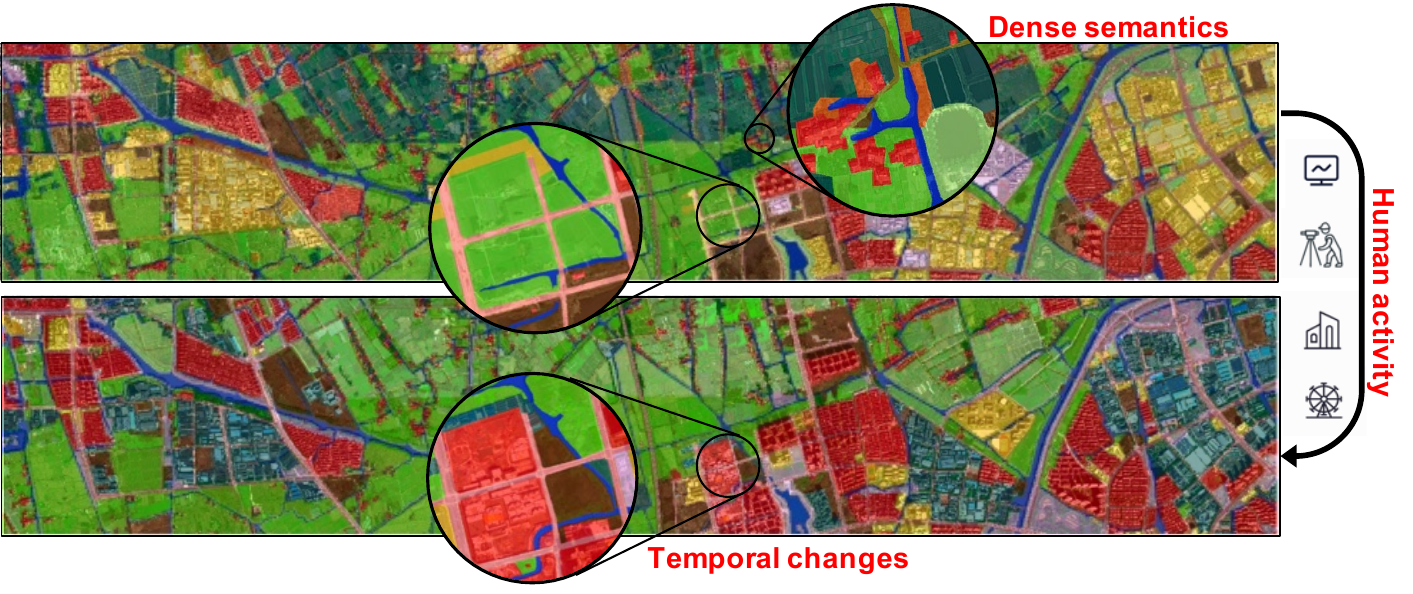}
    \caption{The unique features of urban areas. Compared with other geographic regions, urban areas have dense semantics, fast temporal changes, and involve a large amount of human activities.}
    \label{fig:feature}
\end{figure*}

In the era of data-driven methods, satellite remote sensing provides abundant data for Earth observation and deep learning-based models to comprehend the changes and mechanisms in such observations. However, urban areas have unique features requiring stringent conditions for high-quality data as Fig. \ref{fig:feature} shows. First, multiple semantics are concentrated in small areas, and this dense semantic information is driven by human activities (land use) rather than natural characteristics (land cover) \cite{pacifici2009neural}. This necessitates high-resolution images and fine-grained land use annotations over land cover segmentation datasets. Second, urban areas undergo rapid temporal changes, demanding high-frequency observations to capture these dynamics accurately \cite{perera2023challenges}. Third, Fig. \ref{fig:feature} highlights the diversity of human activities during the urban changes, including work, construction, relocation, and entertainment, requiring multi-source data for effective monitoring. 

Although numerous land cover change segmentation datasets (e.g., LoveDA \cite{wang2021loveda}, SECOND \cite{yang2021asymmetric}, Hi-UCD \cite{tian2020hi}, DynamicEarthNet \cite{toker2022dynamicearthnet}) have been introduced to advance urban monitoring, their coarse-grained land cover classification systems still limit the ability of fine urban semantic understanding. For example, the SECOND dataset only focuses on six classes, including ground, trees, low vegetation, water, buildings, and playgrounds, which fails to capture the full range of urban elements and detailed land use information, thus inadequately reflecting urban conditions and human-urban interactions. Besides, due to the difficulties of acquiring multi-temporal high-resolution images (e.g., cloud obscuration, accessibility), most change segmentation datasets only comprise bi-temporal images with even single-temporal annotations, which cannot match the pace of urban development, leading to challenges in timely planning and management. A high spatial-temporal resolution change segmentation dataset with a fine land use classification system is required.

In this paper, we introduce FUSU, the first multi-temporal, multi-source land use change segmentation dataset with the finest pixel-wise change segmentation annotations to date, covering 17 land use classes and over 30 billion pixels. It includes bi-temporal high-resolution satellite images (0.2-0.5 $m$ resolution) and aligns optical and radar satellite data (Sentinel-2, Sentinel-1) with monthly revisits, enriching temporal and multi-sensor information. Spanning 847 $km^2$ across five major urban districts in northern and southern China, FUSU's geographical diversity ensures domain shifts within the dataset. To leverage this spatial-spectral-temporal-resolution diversity, we propose FUSU-Net, a unified time-series architecture, as a baseline to make full utilization of the enriched information in FUSU for change detection and segmentation tasks. FUSU and FUSU-Net aim to advance dataset and algorithm development for improved urban monitoring and understanding. Our contributions include:
\begin{itemize}
    \item We introduce FUSU, the first land use change segmentation dataset with a fine land use classification system of 17 classes and over 30 billion annotation pixels. FUSU captures timely urban dynamics from different perspectives and bridges the gaps between rich remote sensing data and urban semantic understanding.
    \item We showcase how the constructed time-series data can be leveraged for better urban monitoring by proposing a unified time-series baseline architecture FUSU-Net that conducts end-to-end change detection and segmentation tasks utilizing multi-temporal-source data.
    \item We benchmark FUSU on kinds of methods in several downstream tasks to provide a comprehensive insight.
\end{itemize}


\section{Related Works}

\subsection{Urban Change Segmentation Data}
Urban change segmentation is a critical aspect of Earth observation, garnering significant attention in recent years. Various land cover datasets have been developed to support specific tasks like change detection and segmentation (see Table \ref{tab:data}). ISPRS Potsdam\footnote{https://www.isprs.org/education/benchmarks/UrbanSemLab/2d-sem-label-potsdam.aspx} provides high-resolution images for urban parsing, but it covers small areas and has a limited scale. SpaceNet \cite{van2018spacenet}, EuroSAT \cite{helber2019eurosat}, and GID \cite{GID2020} cover larger areas but suffer from incomplete land cover classification, lower resolution, and single snapshots. LEVIR-CD \cite{chen2020spatial} and WHU \cite{ji2018fully} focus on bi-temporal building change detection, but lack comprehensive semantics. SECOND \cite{yang2021asymmetric}, Hi-UCD \cite{tian2020hi}, and WUSU \cite{shi2023multi} introduce multi-class semantic change detection. However, WUSU and Hi-UCD cover limited regions, and SECOND's coarse annotations and long intervals reduce continual observation capability. LoveDA \cite{wang2021loveda} includes patches from various Chinese cities, but the classification system is coarse-grained, and the annotation only covers a single snapshot time. FLAIR \cite{garioud2024flair} uses aerial and Sentinel-2 images for near-daily observations, yet only one temporal label cannot tell the changes during periods. DynamicEarthNet \cite{toker2022dynamicearthnet} provides daily observations and monthly annotations, but also suffers from the coarse-grained land cover classification system, which fails to provide semantics on human-environment interactions.

In summary, there is a lack of attention to land use datasets. Existing datasets usually present a trade-off among resolution, coverage, snapshot time, annotation pixel, and classification system. On the contrary, FUSU aims for the finest urban semantic understanding, providing the fine-grained land use classification system (17 classes), large-scale annotation pixels (30 billion), high-resolution images (0.2-0.5 m), large coverage (847 $km^2$), temporal information (bi-temporal high-resolution images and monthly Sentinel data), and supporting multiple downstream remote sensing tasks.

\begin{table}[h]
\centering
\caption{A survey on open-source urban change segmentation datasets, including segmentation datasets and change detection datasets.}
\renewcommand{\arraystretch}{1.5}
\resizebox{\linewidth}{!}{
\begin{tabular}{cccccccccccc}
\hline
\multicolumn{2}{c}{\multirow{2}{*}{Dataset}}                             & \multirow{2}{*}{Source}            & \multirow{2}{*}{\begin{tabular}[c]{@{}c@{}}Images\\ (patches)\end{tabular}} & \multirow{2}{*}{Size} & \multirow{2}{*}{\begin{tabular}[c]{@{}c@{}}Area\\ ($km^2$)\end{tabular}} & \multirow{2}{*}{\begin{tabular}[c]{@{}c@{}}Resolution\\ ($m$)\end{tabular}} & \multirow{2}{*}{Class} & \multirow{2}{*}{Objects} & \multirow{2}{*}{\begin{tabular}[c]{@{}c@{}}Temporal\\ (image)\end{tabular}} & \multirow{2}{*}{\begin{tabular}[c]{@{}c@{}}Temporal\\ (annotation)\end{tabular}} & \multirow{2}{*}{\begin{tabular}[c]{@{}c@{}}Ann pixel\\ ($\times 10^9$)\end{tabular}} \\
\multicolumn{2}{c}{}                                                     &                                    &                                                                             &                       &                                                                       &                                                                           &                        &             &                                                                &                                                                           &                                                                            \\ \hline
\multicolumn{1}{c|}{\multirow{6}{*}{\rotatebox{90}{Segmentation}}}     & Potsdam\footnotemark[1]         & Aerial                             & 38                                                                          & 6000                  & 0.05                                                                  & 0.05                                                                      & 6      & \textit{LC}                & 1                                                                           & 1                                                                         & 0.8                                                                        \\
\multicolumn{1}{c|}{}                                  & SpanceNet\cite{van2018spacenet}       & Maxar                              & 60,000                                                                      & 650                   & 5,500                                                                 & 0.3-1.24                                                                  & 2    &  \textit{B\&R}                & 1                                                                           & 1                                                                         & 1.3                                                                        \\
\multicolumn{1}{c|}{}                                  & EuroSAT\cite{helber2019eurosat}         & Sentinel-2                         & 27,000                                                                      & 64                    & 11,059                                                                & 10                                                                        & 10    & \textit{LC}                 & 1                                                                           & 1                                                                         & 0.1                                                                        \\
\multicolumn{1}{c|}{}                                  & GID\cite{GID2020}             & Gaofen-2                           & 150                                                                         & 6800-7200             & 50,000                                                                & 1                                                                         & 5/15     & \textit{LC}              & 1                                                                           & 1                                                                         & 7.3                                                                        \\
\multicolumn{1}{c|}{}                                  & LoveDA\cite{wang2021loveda}          & Google Earth                       & 5987                                                                        & 1024                  & 536                                                                   & 0.3                                                                       & 6    & \textit{LC}                  & 1                                                                           & 1                                                                         & 6.3                                                                        \\
\multicolumn{1}{c|}{}                                  & FLAIR\cite{garioud2024flair}           & Aerial/Sentinel-2                  & 77,762                                                                      & 512/40                & 817                                                                   & 0.2/10                                                                    & 18    & \textit{LC}                 & 4 days                                                                      & 1                                                                         & 20.3                                                                       \\ \hline
\multicolumn{1}{c|}{\multirow{6}{*}{\rotatebox{90}{Change Detection}}} & LEVIR-CD\cite{chen2020spatial}        & Google Earth                       & 637                                                                         & 1024                  & 167                                                                   & 0.5                                                                       & 1    & \textit{B}                  & 2                                                                           & 1                                                                         & 0.005                                                                      \\
\multicolumn{1}{c|}{}                                  & WHU\cite{ji2018fully}             & Aerial                             & 8,189                                                                       & 512                   & 192                                                                   & 0.3                                                                       & 1    & \textit{B}                  & 2                                                                           & 1                                                                         & 0.4                                                                        \\
\multicolumn{1}{c|}{}                                  & SECOND\cite{yang2021asymmetric}          & Satellite                          & 4,662                                                                       & 512                   & 1,200                                                                  & 0.5-1                                                                     & 6 & \textit{LC}                     & 2                                                                           & 2                                                                         & 0.9                                                                        \\
\multicolumn{1}{c|}{}                                  & Hi-UCD\cite{tian2020hi}          & Aerial                             & 1,293                                                                       & 1024                  & 30                                                                    & 0.1                                                                       & 9     & \textit{LC}                 & 3                                                                           & 3                                                                         & 2.7                                                                        \\
\multicolumn{1}{c|}{}                                  & WUSU\cite{shi2023multi} & Gaofen-2                           & 2                                                                           & 5500-7025             & 80                                                                    & 1                                                                         & 11   & \textit{LC}                  & 3                                                                           & 3                                                                         & 1.5                                                                        \\
\multicolumn{1}{c|}{}                                  & DynamicEarthNet\cite{toker2022dynamicearthnet} & PlanetFusion                       & 54,750                                                                      & 1024                  & 16,986                                                                & 3                                                                         & 7      & \textit{LC}                & daily                                                                       & monthly                                                                   & 1.9                                                                        \\ \hline
\rowcolor[HTML]{FAE3E1} 
\multicolumn{1}{l}{}                                   & \textbf{FUSU}   & \textbf{Google Earth/Sentinel-1/2} & \textbf{62,752}& \textbf{512/128}      & \textbf{847}                                                          & \textbf{0.2-0.5/10}                                                          & \textbf{17}   & \textbf{\textit{LU}}         & \textbf{monthly}                                                            & \textbf{2}                                                                & \textbf{32.2}                                                              \\ \hline
\multicolumn{4}{l}{\small $\bullet$ \textit{B}-Buildings, \textit{R}-Roads, \textit{LC}-Land Cover, \textit{LU}-Land Use.}\\
\end{tabular}
}
\label{tab:data}
\end{table}




\subsection{Remote sensing tasks}

\textbf{Change Detection} identifies surface differences by processing images of the same area captured at different times \cite{shi2020change}. It includes binary change detection \cite{chen2020spatial,ji2018fully}, which detects changes in a single class (changed or unchanged), and semantic change detection \cite{yang2021asymmetric,toker2022dynamicearthnet}, which provides detailed land semantics. High-frequency observations are essential for timely geographical change detection, and fine-grained annotations improve precision. However, most datasets provide only bi-temporal observations due to the challenge of acquiring high-resolution multi-temporal images, resulting in long intervals that impede timely monitoring. Additionally, the lack of fine-grained multi-temporal annotations restricts the development of semantic change detection algorithms.

These challenges highlight the need for richer temporal data and fine-grained land use classifications, as well as methods capable of handling multi-temporal information. Current datasets' coarse-grained classifications do not accurately reflect urban conditions, and integrating multi-temporal data from other accessible sensors to enhance change detection has been underexplored. To address these issues, we propose FUSU, which includes bi-temporal fine-grained annotations and multi-temporal observations from high-resolution and Sentinel images. We also design a new unified architecture FUSU-Net to leverage time-series information for semantic change detection and segmentation.

\textbf{Semantic segmentation} has been widely applied in remote sensing for tasks such as land cover mapping \cite{dong2023large}, building/road extraction \cite{yuan2020long,zhang2024swcare}, and cropland cover mapping \cite{garnot2021panoptic}. Encoder-decoder architectures are well-suited to the diverse nature of remote sensing images \cite{toker2022dynamicearthnet}. Most studies focus on segmenting objects from static images \cite{robinson2019large,wang2021loveda}, while some have used time-series images to improve performance \cite{garnot2021panoptic,tarasiou2023vits}. Our FUSU-Net integrates time-series information into the bi-temporal segmentation task. We believe the unique time-series structure of FUSU will inspire the development of more advanced time-series segmentation algorithms in remote sensing.

\section{FUSU Dataset}\label{sec:fusu}

\begin{figure*}[ht]
    \centering
    \includegraphics[width=0.9\linewidth]{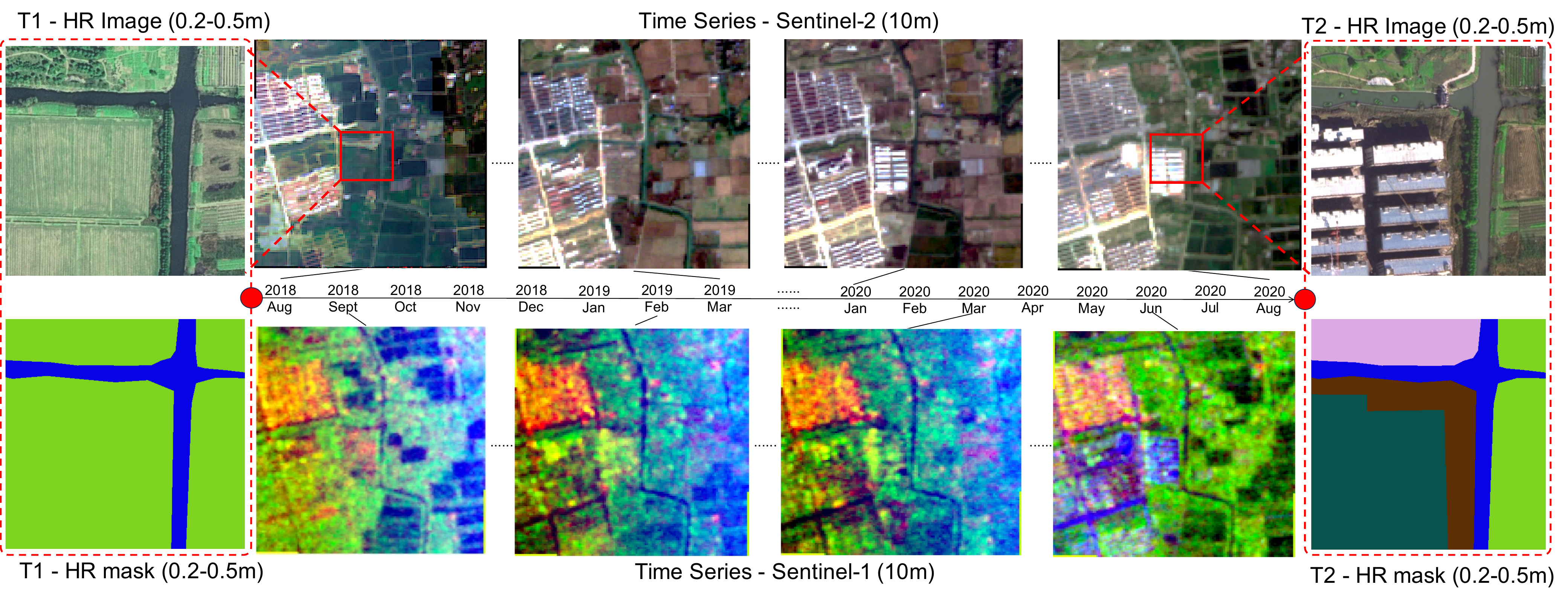}
    \caption{The visualization of the FUSU dataset construction. Each patch has 27 images (25 Sentinel images and 2 high-resolution images), and 2 labels. The content of the high-resolution image is center-surrounded by the Sentinel image as the red rectangle shows.}
    \label{fig:overall}
\end{figure*}

We introduce FUSU, a multi-temporal, multi-source change segmentation dataset for fine-grained urban semantic understanding. FUSU consists of 62,752 image patches, each containing 27 images from three sources with different resolutions and snapshot times, and includes two annotations as shown in Fig. \ref{fig:overall}. FUSU has four key features:

\textbf{Fine-grained:} FUSU features the finest land use classification system in change segmentation datasets, with bi-temporal dense annotations. It includes 17 classes—artificial-constructed, agricultural, and natural—that detail urban functional zoning and enhance understanding of urban structural development.

\textbf{Multi-temporal:} FUSU offers time-series observations with monthly revisits. Along with bi-temporal high-resolution images and fine-grained annotations, it supports high-frequency urban monitoring, enabling methods to leverage long-range temporal context for better inferences.

\textbf{Multi-source:} FUSU combines data from three satellite sources (Google Earth, Sentinel-2, Sentinel-1) with different temporal, resolution, and band compositions. Each image patch unifies spatial, temporal, and spectral contexts, providing richer information than single-source data.

\textbf{Domain shifts:} FUSU covers five urban areas in northern and southern China, each with diverse geographical features and urban landscapes. Variability in climate types and class ratios across these regions contribute to representation gaps and pronounced domain shifts in the feature data.

\subsection{Construction of FUSU}

\textbf{Acquisitions.} FUSU uses three data sources with different resolutions, geographical details, and acquisition times. Google Earth images are $512\times512$ pixels with a 0.3 m resolution and RGB bands. Sentinel-1 and Sentinel-2 images are sourced from Google Earth Engine (GEE). Sentinel-1 images are preprocessed by GEE (noise removal, radiometric calibration, orthorectification). Sentinel-2 images undergo cloud removal, atmospheric correction, radiometric calibration, and orthorectification, then are concatenated with Sentinel-1 data. Each Sentinel image is $128\times128$ pixels with a 10 m resolution and 14 bands. Google Earth and Sentinel patches are not strictly aligned; Google Earth patches cover only the central area of corresponding Sentinel patches (Fig. \ref{fig:overall}). This approach preserves semantic detail and captures broader context, aiding spatial dynamics understanding. More details are in the Sec. \ref{sec:expand}.

\begin{figure}[t]
    \centering
    \includegraphics[width=0.95\linewidth]{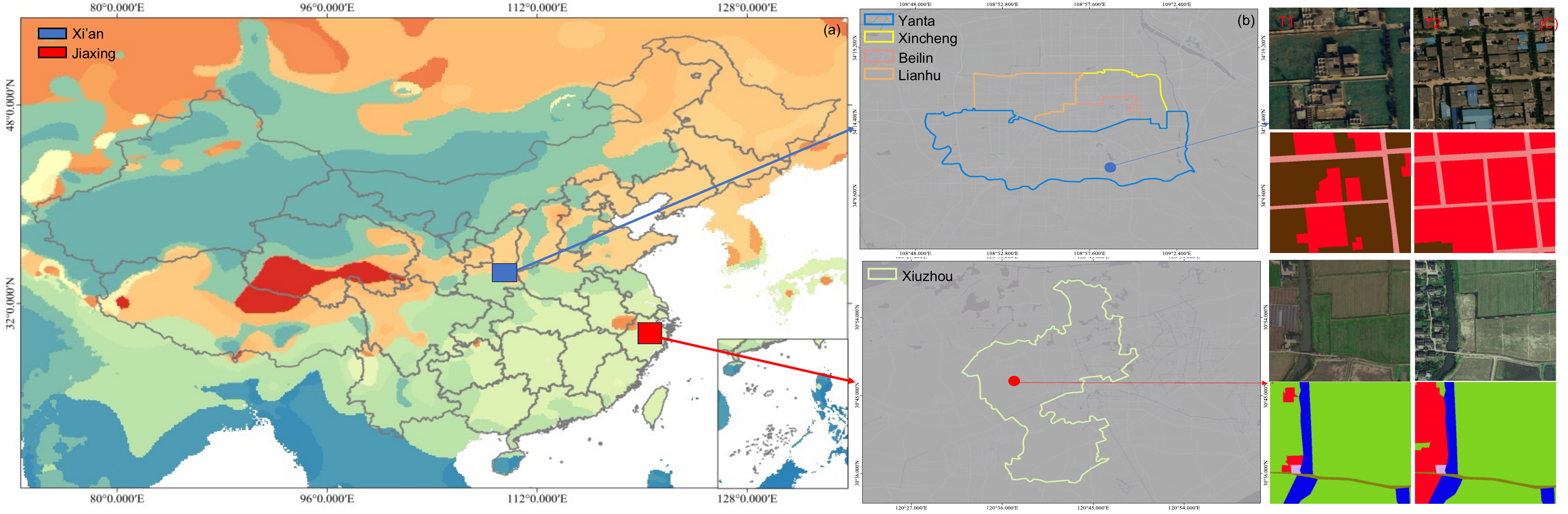}
    \caption{The distribution of the FUSU dataset. (a) Xi'an and Jiaxing are located in different climate zones. (b) The 5 urban districts of Xi'an and Jiaxing in FUSU dataset. (c) The visualization of image samples.}
    \label{fig:distribution}
\end{figure}

\textbf{Distribution.} FUSU covers 847 $km^2$ across five urban districts in China: Xiuzhou in Jiaxing, and Yanta, Beilin, Xincheng, and Lianhu in Xi'an. The different climates of Jiaxing and Xi'an are illustrated in Fig. \ref{fig:distribution}(a). FUSU provides continuous monthly observations from August 2018 to August 2020. Google Earth images were captured in August 2018 and August 2020, while Sentinel-1 and Sentinel-2 images were collected monthly between these dates.

\textbf{Annotations.} Bi-temporal Google Earth images are manually annotated pixel-wise by two teams of geography experts. Table \ref{tab:class} shows the classes, label values and colors. More details about the annotations can be found in Sec. \ref{sec:ann}.

\begin{table}[h]
\centering
\caption{Land use classification system of FUSU and corresponding label values, colors.}
\resizebox{0.9\linewidth}{!}{
\begin{tabular}{ccccccccc}
\hline
Color                    & Class                    & Label Value & Color                    & Class               & Label Value & Color                    & Class        & Label Value \\ \hline
\cellcolor[HTML]{E98585} & Traffic land             & 1           & \cellcolor[HTML]{0A524D} & Industrial land     & 7           & \cellcolor[HTML]{F3E5B0} & Special land & 13          \\
\cellcolor[HTML]{089AE6} & Inland water             & 2           & \cellcolor[HTML]{B8E986} & Orchard                & 8           & \cellcolor[HTML]{036400} & Forest       & 14          \\
\cellcolor[HTML]{FF001E} & Residential land         & 3           & \cellcolor[HTML]{DBAAE6} & Park           & 9           & \cellcolor[HTML]{7F7B7F} & Storage      & 15          \\
\cellcolor[HTML]{7ED321} & Cropland                 & 4           & \cellcolor[HTML]{FFC702} & Public management   & 10          & \cellcolor[HTML]{34CDF9} & wetland      & 16          \\
\cellcolor[HTML]{877E14} & Agriculture construction & 5           & \cellcolor[HTML]{FCE805} & Commercial land     & 11          & \cellcolor[HTML]{12E3B4} & Grass        & 17          \\
\cellcolor[HTML]{5E2F04} & Blank                    & 6           & \cellcolor[HTML]{F56B00} & Public construction & 12          & \cellcolor[HTML]{FFFFFF} & Background   & 0           \\ \hline
\end{tabular}
}
\label{tab:class}
\end{table}

\begin{figure}[htp]
    \centering
    \includegraphics[width=0.95\linewidth]{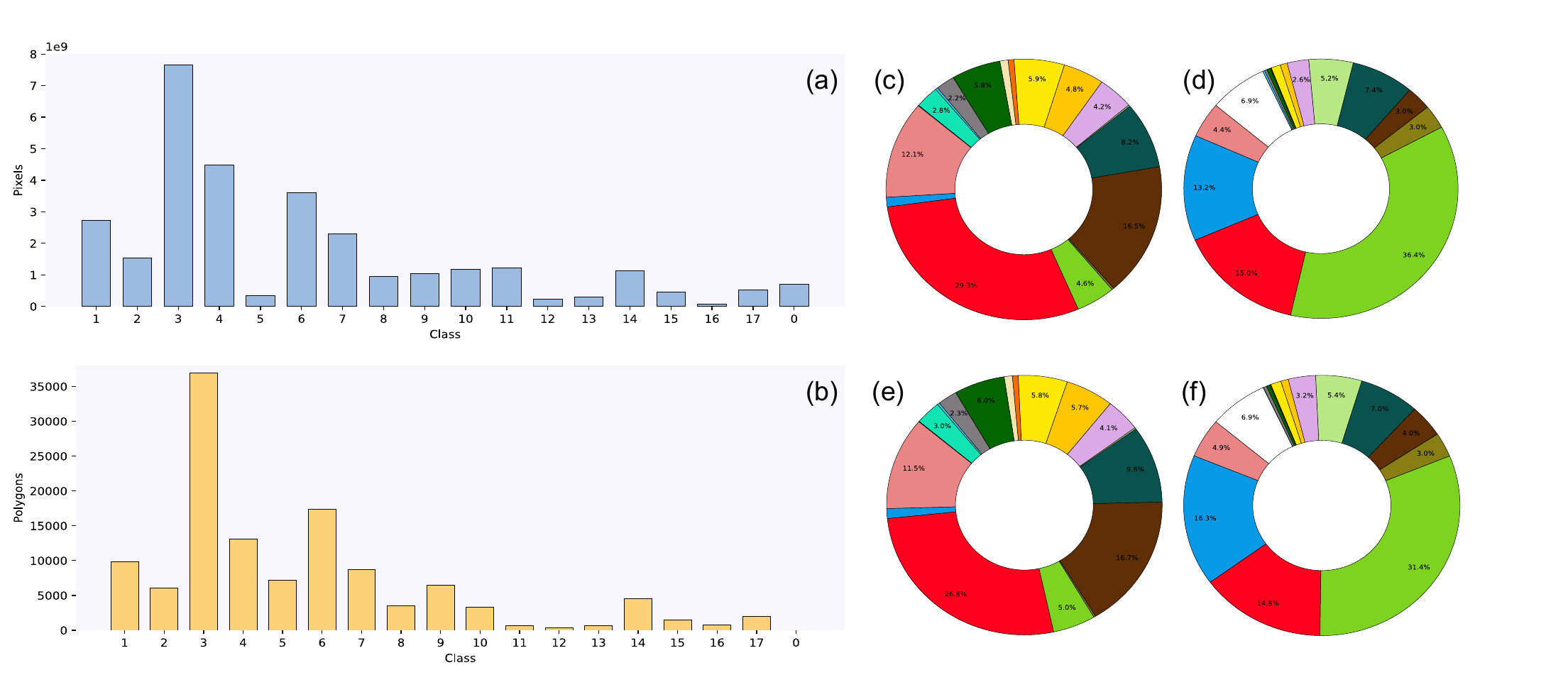}
    \caption{The statistic of the FUSU dataset. (a) Pixels distribution. (b) Polygon distribution. (c) Class distribution of T1 Xi'an. (d) Class distribution of T1 Jiaxing. (e) Class distribution of T2 Xi'an. (f) Class distribution of T2 Jiaxing.} 
    \label{fig:statistic}
\end{figure}

\subsection{Statistic}
FUSU includes bi-temporal pixel-level annotations covering 17 land use classes. Fig. \ref{fig:statistic}(a) and (b) illustrate the distribution of pixels and polygons for each class at a single time snapshot. Residential land dominates both in terms of polygons and pixels. Some classes, like agriculture construction land, exhibit asymmetrical distributions. The highly unbalanced distribution numbers show a ratio exceeding 90 between the most and least frequent types. Fig. \ref{fig:statistic}(c)-(f) display the class ratios in Xi'an and Jiaxing at two-time snapshots, revealing varying distributions between the cities. Jiaxing is characterized by significant cropland and residential areas, while Xi'an has more commercial land. These class imbalances and city differences pose challenges for urban monitoring using FUSU.

\section{FUSU-Net} \label{sec:fusunet}

To fully utilize FUSU, we propose a unified time-series baseline architecture named FUSU-Net that conducts end-to-end change detection and segmentation tasks. Fig. \ref{fig:fusunet} shows the architecture.

\subsection{Preliminary and Overview} \label{sec:pre}

Given T1 image $\mathcal{I}_1$, T2 image $\mathcal{I}_2$, the corresponding groundtruth labels $\mathcal{Y}_1$, $\mathcal{Y}_2$, and the time-series temporal images $\mathcal{I}_T$, we have two ultimate goals: build a segmentation function $\mathcal{F}_s$ that generates segmentation map $\hat{\mathcal{Y}} = \mathcal{F}_s(\mathcal{I}\ |\ \mathcal{I}_T)$, and build a change detection function $\mathcal{F}_c$ that find binary changes between two input images $\hat{\mathcal{C}} = \mathcal{F}_c(\hat{\mathcal{Y}_1}, \hat{\mathcal{Y}_2}\ |\ \mathcal{I}_T)$. These two goals mean we need to optimize the loss $\mathcal{L}$ between predicted values and labels:
\begin{equation}
    \mathcal{\theta^{*}} = \mathop{\arg\min}_{\theta}\{\mathcal{L}^s(\mathcal{F}_s(\mathcal{I}\ |\ \mathcal{I}_T ), \mathcal{Y}) + \mathcal{L}^c(\mathcal{F}_c(\hat{\mathcal{Y}_1}, \hat{\mathcal{Y}_2}\ |\ \mathcal{I}_T), \mathcal{Y}_c)\}, \label{1}
\end{equation}
where $\mathcal{\theta^{*}}$ is the optimized learned parameters generated by the optimized $\mathcal{L}^c$ and $\mathcal{L}^s$, and $\mathcal{\theta}$ represents the learned parameters, and $\mathcal{Y}_c$ is the binary change groundtruth label, which can be generated by $\mathcal{Y}_1$, $\mathcal{Y}_2$:
\begin{equation}
y_c^{(i,j)} = 
\left\{
    \begin{array}{lr}
    0, & y_1^{(i,j)}\ =\ y_2^{(i,j)}\\
    1, & y_1^{(i,j)}\ \ne\ y_2^{(i,j)},\\  
    \end{array}
\right.
\label{2}
\end{equation}

where $y^{(i,j)}$ is the pixel value. Assuming the additional temporal and spectral information in time-series images can guide the high-resolution segmentation and change detection, we further extract the high-level temporal and spectral information and use $\mathcal{Y}_1$ for supervision. Thus the optimization body can be divided into:

\begin{align}
\mathcal{\theta^{*}} = &\mathop{\arg\min}_{\theta}\{\mathcal{L}_1^s(\mathcal{F}_s(\mathcal{I}\ |\ \mathcal{F}_s(\mathcal{I}_T, \mathcal{Y}_1; {\theta}), \mathcal{Y}_1; {\theta}) + \mathcal{L}_2^s(\mathcal{F}_s(\mathcal{I}\ |\ \mathcal{F}_s(\mathcal{I}_T, \mathcal{Y}_1; {\theta}), \mathcal{Y}_2; {\theta}) + 
\nonumber \\
&\mathcal{L}_T^s(\mathcal{F}_s(\mathcal{I}_T, \mathcal{Y}_1; {\theta})+\mathcal{L}^c(\mathcal{F}_c(\hat{\mathcal{Y}_1}, \hat{\mathcal{Y}_2}\ |\ \mathcal{F}_s(\mathcal{I}_T, \mathcal{Y}_1; {\theta}), \mathcal{Y}_c; {\theta})\},      
\end{align}
where $\mathcal{L}_{\{1,2,T\}}^s$ is the loss of segmentation of T1 image, T2 image, and time-series images, respectively. 

\subsection{Overall architecture}

\begin{figure*}[t]
    \centering
    \includegraphics[width=0.95\linewidth]{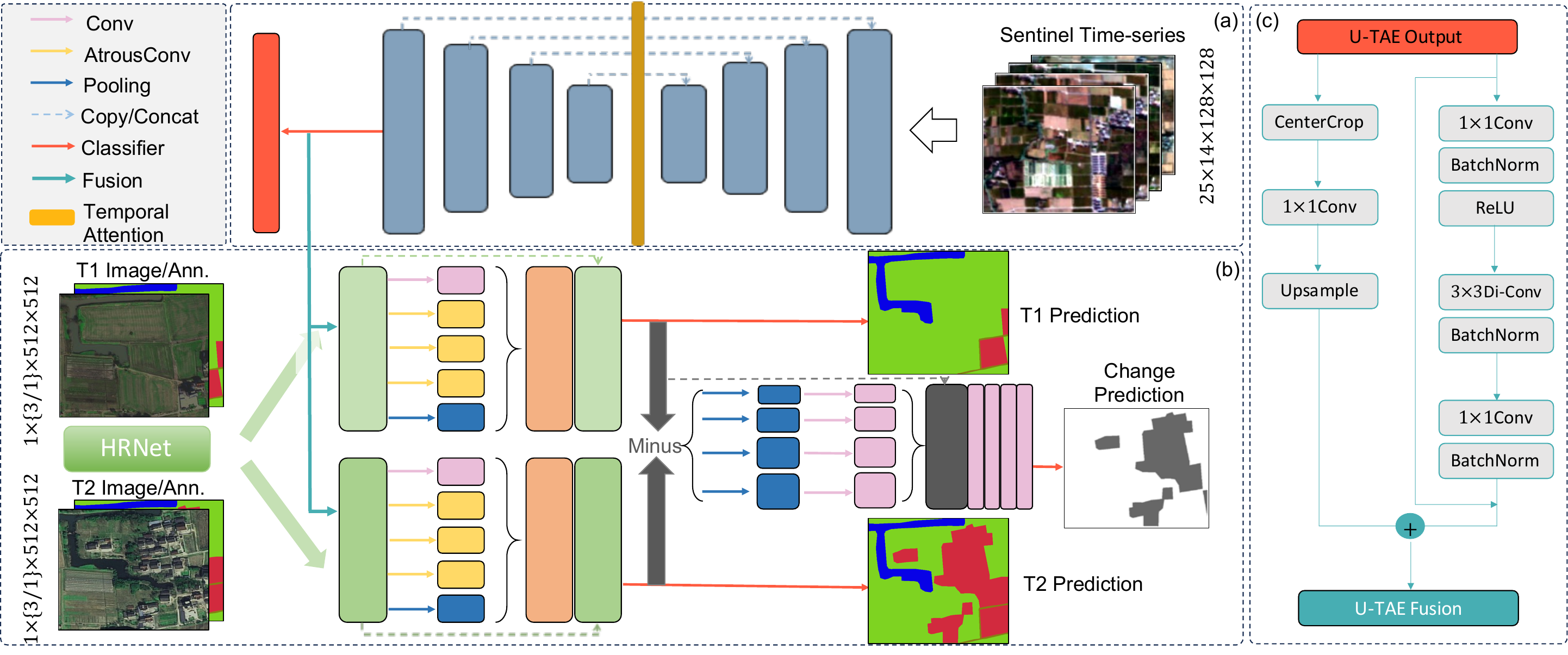}
    \caption{The architecture of FUSU-Net. (a) U-TAE branch for time-series images. (b) Bi-temporal branch for segmentation and change detection. (c) Feature fusion.}
    \label{fig:fusunet}
\end{figure*}

As Fig. \ref{fig:fusunet} shows, the overall architecture of FUSU-Net includes two branches: (a) This branch processes Sentinel time-series images and outputs time-series features; (b) This branch processes bi-temporal high-resolution images and annotations and outputs both bi-temporal segmentation results and change detection results. 

As Fig. \ref{fig:fusunet}(a) shows, to process the Sentinel time-series images, we use U-TAE \cite{garnot2021panoptic} with temporal attention to effectively capture temporal information in feature maps at various resolutions. The input shape is $25\times14\times512\times512$ ($T\times C \times H \times W$) and the output shape is $64\times512\times512$. Fig. \ref{fig:fusunet}(b) shows that we first use an HR-Net pre-trained on ImageNet as the backbone to extract bi-temporal features. Then we input each feature into separated ASPP \cite{chen2017rethinking} segmentation heads to get the segmentation results. We then conduct a minus operation between bi-temporal segmentation features, and after a Spatial Pyramid Pooling head \cite{zhao2017pyramid}, we can get the binary change detection result. Note that Fig. \ref{fig:fusunet}(c) shows the fusion module. Time-series features fuse with bi-temporal features via two transformations: First, the time-series feature is center-cropped to strictly geographically align with the bi-temporal features. Then after a $1\times1$ convolution and upsampling layer, the center-cropped feature has the same shape with bi-temporal features. Second, we reserve the large spatial information of the time-series feature and after a bottle-neck structure with a dilated convolution, we map it to the same shape of the bi-temporal features. An add operation is conducted for the feature fusion.

\subsection{Loss Functions}

As discussed in Sec. \ref{sec:pre}, we use 4 loss functions to train FUSU-Net: three segmentation loss $\mathcal{L}_{\{1,2,T\}}^s$, and a change loss $\mathcal{L}^c$. The segmentation loss functions are the multi-class cross-entropy loss. Specifically, for time-series supervision, we first centercrop the output for geographical alignment, then upsample it to the same size of groundtruth label $\mathcal{Y}_1$. The change loss is the BCE loss to supervise the binary changes. More details about supervision and implementation can be found in Sec. \ref{sec:supervision}.

\section{Experiments}
We utilize our dataset for semantic segmentation in Sec. \ref{sec:seg} and change detection in Sec. \ref{sec:cd} with various experiments on state-of-the-art baseline methods and FUSU-Net. We also validate the feature disparities between Jiaxing and Xi'an in the segmentation task.

\subsection{Semantic Segmentation} \label{sec:seg}
Land use segmentation is crucial for urban monitoring. We focus on single-temporal images and labels for this semantic segmentation task. We compare other seven baseline segmentation methods with our FUSU-Net: FCN \cite{long2015fully}, PSPNet \cite{zhao2017pyramid}, Fast-SCNN \cite{poudel2019fast}, Deeplab-v3 \cite{chen2017rethinking}, HRNet \cite{wang2020deep}, K-net \cite{zhang2021k}, and U-TAE \cite{garnot2021panoptic}. Evaluation is based on intersection over union (IoU) per class and mean IoU (mIoU) across all 17 land use classes, following established protocols. Additionally, we investigate feature disparities between Jiaxing and Xi'an through two experiments: intra-dataset (whole, Xi'an, Jiaxing) and inter-dataset (training on one, testing on the other). Implementation details are provided in the Sec. \ref{sec:detail}.

\begin{table}[h]
\centering
\caption{Semantic segmentation results obtained from intra-dataset.}
\renewcommand{\arraystretch}{1.5}
\resizebox{\linewidth}{!}{
\begin{tabular}{ccccccccccccccccccc}
\hline
\multirow{2}{*}{Method} & \multicolumn{17}{c}{IoU per class (\%)}                                                                                                & \multirow{2}{*}{mIoU} \\ \cline{2-18}
                        & 1     & 2     & 3     & 4      & 5     & 6     & 7     & 8     & 9     & 10    & 11    & 12    & 13    & 14    & 15    & 16    & 17    &                       \\ \hline
FCN \cite{long2015fully}                  & 70.83  & 76.49 & 74.67  & 84.14  & 30.84 & \textbf{52.16} & 53.39 & 33.32 & \textbf{52.7} & \textbf{50.24} & 28.98 & 0.09 & 30.62 & 57.42 & 23.61 & 13.04 & 17.79 & 44.25                \\
PSPNet \cite{zhao2017pyramid}                 & 65.37 & 79.15 & 71.44 & 82.354 & 23.5  & 49.97 & 52.82 & 40.44 & 44.90  & 44.50  & 31.39 & 30.08 & 24.04 & 48.69 & 41.64 & 24.50  & 32.58 & 46.32                 \\
Fast-SCNN \cite{poudel2019fast}              & 54.42 & 72.28 & 66.25 & 78.97  & 2.80   & 42.40  & 47.84 & 35.33 & 30.24 & 30.24 & 31.94 & 12.03 & 0     & 0     & \textbf{44.66} & 31.75 & 23.39 & 35.56                 \\
Deeplab-v3 \cite{chen2017rethinking}             & 66.17 & 77.31 & 71.20  & 82.10   & 26.30  & 49.61 & 53.96 & 37.35 & 45.85 & 47.61 & 33.21 & \textbf{35.06} & \textbf{30.68} & 54.14 & 34.94 & \textbf{34.15} & 32.07 & 47.74                 \\
HRNet \cite{wang2020deep}                 & 67.6  & \textbf{80.39} & 73.24 & 83.02  & 22.94 & 49.00  & 54.05 & 40.10  & 46.43 & 49.12 & 31.66 & 26.68 & 15.21 & 52.38 & 42.84 & 30.16 & 32.09 & 46.88                 \\
K-net \cite{zhang2021k}                    & 59.97 & 72.68 & 66.87 & 79.46  & 18.45 & 44.19 & 48.07 & 32.05 & 35.07 & 35.64 & 19.9  & 18.44 & 18.69 & 49.61 & 29.2  & 23.88 & 22.51 & 39.69                 \\
U-TAE \cite{garnot2021panoptic}                    & 59.57 & 64.18 & 65.76 & 77.92  & 24.87 & 40.13 & 46.75 & 29.89 & 41.72 & 30.57 & 26.13  & 6.85 & 25.96 & 30.57 & 41.83  & 15.08 & 8.12 & 37.63                 \\
\hline \rowcolor[HTML]{FAE3E1} 
\textbf{FUSU-Net}                &   \textbf{74.79}    &   78.95    &   \textbf{76.13}    &  \textbf{85.35}      &   \textbf{34.81}    &   50.54    &    \textbf{53.47}   &   \textbf{41.50}    &  49.64     &   45.78    &   \textbf{36.69}    &  28.85     &    28.98   &   \textbf{60.21}    &   44.41    &    30.07   &    \textbf{33.69}   &    \textbf{50.10}                   \\ \hline
\end{tabular}}
\label{tab:seg}
\end{table}

\begin{table}[h]
\centering
\caption{Semantic segmentation results obtained from inter-dataset.}
\renewcommand{\arraystretch}{1.5}
\resizebox{0.62\linewidth}{!}{
\begin{tabular}{ccccc}
\hline
\multirow{2}{*}{Method} & \multicolumn{4}{c}{mIoU}                                                                                                                                                                                                                                                                                                          \\ \cline{2-5} 
                        & \begin{tabular}[c]{@{}c@{}}Training on Xi'an\\ Testing on Xi'an\end{tabular} & \begin{tabular}[c]{@{}c@{}}Training on Jiaxing\\ Testing on Jiaxing\end{tabular} & \begin{tabular}[c]{@{}c@{}}Training on Xi'an\\ Testing on Jiaxing\end{tabular} & \begin{tabular}[c]{@{}c@{}}Training on jiaxing\\ Testing on Xi'an\end{tabular} \\ \hline
FCN \cite{long2015fully}                     & 50.21                                                                        & 45.53                                                                            & 9.07                                                                           & 9.36                                                                           \\
PSPNet \cite{zhao2017pyramid}                 & 46.52                                                                        & 43.35                                                                            & 8.55                                                                           & 9.72                                                                           \\
Fast SCNN \cite{poudel2019fast}              & 32.97                                                                        & 32.76                                                                            & 7.83                                                                           & 8.51                                                                           \\
HRNet \cite{wang2020deep}                  & 46.78                                                                        & 45.01                                                                            & 10.07                                                                          & 9.73                                                                           \\
K-net \cite{zhang2021k}                    & 38.17                                                                        & 37.41                                                                            & 9.31                                                                           & \textbf{10.59}                                                                          \\
Deeplab-v3 \cite{chen2017rethinking}                  & 47.30                                                                        & 47.89                                                                            & 9.65                                                                           & 9.13                                                                           \\ \hline \rowcolor[HTML]{FAE3E1}
\textbf{FUSU-Net}                & \textbf{53.63}                                                                        & \textbf{49.91}                                                                            & \textbf{11.65}                                                                          & 10.46                                                                          \\ \hline
\end{tabular}
}
\label{tab:crossdata}
\end{table}

\textbf{Overall results.} Table \ref{tab:seg} shows the segmentation results. We observe that FUSU-Net achieves the best results regarding mIoU. Specifically, FUSU-Net performs better than other methods not only on some comparatively simple classes (i.e., traffic land-1, residential land-3) but also has continuous promising results on difficult classes where other methods have poor performance (i.e., commercial land-11, special land-13). Note that FUSU-Net is backboned by HRNet and the segmentation head is PSPNet with FCN, and the results directly show the benefits of adding features of time-series Sentinel images. When compared with U-TAE, we can see that high-resolution images can also improve performance by providing more clear observation details.

\textbf{Cross-dataset results.} Table \ref{tab:crossdata} shows the segmentation results with different training and testing datasets. There is a dramatic drop in mIoU on cross-dataset training and testing compared with training and testing on the same datasets. We can tell the huge feature differences between Jiaxing and Xi'an from these results.

\begin{table}[h]
\caption{Semantic change detection results obtained from intra-dataset.}
\renewcommand{\arraystretch}{1.5}
\resizebox{\linewidth}{!}{
\begin{tabular}{ccccccccccccccccccc}
\hline
                         & \multicolumn{17}{c}{IoU per class (\%)}                                                                                             &                        \\ \cline{2-18}
\multirow{-2}{*}{Method} & 1     & 2     & 3     & 4     & 5     & 6     & 7     & 8     & 9     & 10    & 11    & 12   & 13   & 14    & 15    & 16    & 17    & \multirow{-2}{*}{mIoU} \\ \hline
BIT \cite{chen2021remote}                     & 35.54 & 48.90 & 46.89 & 42.27 & 4.01  & 46.70 & 59.92 & 23.6  & 35.41 & 25.82 & 17.88 & 0    & 3.95 & 54.23 & 22.62 & 12.20 & 46.15 & 30.95                  \\
ChangeFormer \cite{bandara2022transformer}            & 39.31 & 57.87 & 57.13 & 39.42 & 9.20  & 25.58 & 60.11 & 31.33 & 27.17 & 19.79 & 12.07 & 0.31 & 7.42 & 59.81 & 19.71 & \textbf{35.61} & 35.13 & 32.17                  \\
ICIFNet \cite{feng2022icif}                 & 49.75 & 56.41 & 62.23 & 51.21 & 4.7   & 53.81 & \textbf{61.43} & 30.03 & \textbf{47.35} & 3.47  & 10.45 & 0    & 0    & \textbf{73.65} & 53.18 & 11.15 & 60.75 & 36.17                  \\
DMINet  \cite{feng2023change} & 26.63 & 34.08 & 54.91 & 42.75 & 0     & 32.59 & 39.80 & 17.91 & 19.34 & 0     & 6.35  & 0    & 0    & 21.83 & 39.41 & 20.94 & 54.18 & 24.16                  \\
SSCD-l \cite{ding2022bi}                  & 23.19 & 15.95 & 31.32 & 29.12 & 6.12  & 35.46 & 27.08 & 12.30 & 18.91 & 2.50  & 0     & 0    & 3.39 & 2.06  & 20.51 & 16.34 & 15.69 & 15.29                  \\
Bi-SRNet \cite{ding2022bi}                & 26.19 & 41.42 & 39.82 & 40.01 & \textbf{21.18} & 44.26 & 46.59 & 26.70 & 25.05 & \textbf{31.23} & \textbf{20.21} & \textbf{7.18} & 4.74 & 40.91 & 31.66 & 30.87 & 37.40 & 30.91                  \\ \hline
\rowcolor[HTML]{FAE3E1} 
\textbf{FUSU-Net}                 & \textbf{55.67} & \textbf{61.46} & \textbf{66.19} & \textbf{55.83} & 19.82   & \textbf{55.22} & 57.86 & \textbf{34.59} & 46.43 & 15.45  & 16.31 & 5.89    & \textbf{9.47}    & 65.12 & \textbf{54.32} & 14.45 & \textbf{64.52} & \textbf{41.09}                      \\ \hline

\end{tabular}}
\label{tab:CD}
\end{table}

\subsection{Change Detection} \label{sec:cd}
We then compare the performance of change detection baselines on FUSU. Here, we complete the binary change detection (BCD) experiment and semantic change detection (SCD) experiment. For binary change detection, we introduce 6 methods: DMINet \cite{feng2023change}, ICIFNet \cite{feng2022icif}, ChangeFormer \cite{bandara2022transformer}, A2Net \cite{li2023lightweight}, BIT \cite{chen2021remote}, USSFC-Net \cite{lei2023ultralightweight}. We evaluate the results by IoU on changed pixels. For semantic change detection, we introduce 6 methods: BIT \cite{chen2021remote}, ChangeFormer \cite{bandara2022transformer}, ICIFNet \cite{feng2022icif}, DMINet \cite{feng2023change}, SSCD-l \cite{ding2022bi}, Bi-SRNet \cite{ding2022bi}. We evaluate the change detection results by IoU per class and mIoU over all 17 land use classes. Implementation details can be found in Sec. \ref{sec:detail}.

\begin{wraptable}{r}{6cm}
\centering
\renewcommand{\arraystretch}{1.5}
\scriptsize
\caption{Binary change detection results obtained from intra-dataset.}
\resizebox{0.9\linewidth}{!}{
\begin{tabular}{cccc}
\hline
Method & IoU & Method & IoU\\ \hline
BIT \cite{chen2021remote}  & 47.91 & ChangeFormer \cite{bandara2022transformer}& 59.64\\
ICIFNet \cite{feng2022icif}                & 64.74                  &
DMINet \cite{feng2023change}                   & 72.59                  \\
A2Net \cite{li2023lightweight}                 & 69.22                &
USSFC-Net \cite{lei2023ultralightweight}           & 62.85                \\ \hline
\rowcolor[HTML]{FAE3E1} 
FUSU-Net             & 79.80 \\
\hline
\end{tabular}}
\label{tab:bcd}
\end{wraptable}

\textbf{Overall results.} Table \ref{tab:bcd} and Table \ref{tab:CD} present the results of binary and semantic change detection. In binary change detection, with only unchanged and changed pixels, class-specific IoU is not applicable. Our FUSU-Net outperforms other baselines by 7.21\%-31.89\% in IoU. In semantic change detection, challenging classes such as public management-10, public construction-12, and special land-13 are observed across all methods, consistent with semantic segmentation results. Notably, FUSU-Net achieves better performance compared to other baseline methods than it does in the semantic segmentation task, which can be attributed to continuous observation and change information provided by time-series Sentinel images between two high-resolution image snapshots.

\section{Discussion} \label{sec:dis}

\begin{wraptable}{l}{5cm}
\centering
\renewcommand{\arraystretch}{1.5}
\scriptsize
\caption{Ablation results on the effectiveness of time-series.}
\resizebox{0.9\linewidth}{!}{
\begin{tabular}{ccccc}
\hline
Time-series & 0 & 9 & 18 & 25 \\
\hline
mIoU (Seg) & 46.72 & 47.19 & 48.47 & 50.10\\
IoU (BCD) & 65.51 & 69.35 & 74.39 & 79.80\\
mIoU (SCD) & 26.64 & 34.14 & 36.55 & 41.09\\
\hline
\end{tabular}}
\label{tab:abl}
\end{wraptable}

\textbf{Effectiveness of time-series.} We evaluate to what extent time-series images enhance the performance. Table \ref{tab:abl} shows the results. We choose the number of time-series images as the variable (i.e., all time-series images, partial time-series images, zero time-series images). We can see for the FUSU dataset, more time-series images contribute to better results. It is desirable to use all time-series images as additional temporal information. 

\begin{wraptable}{r}{5cm}
\centering
\renewcommand{\arraystretch}{1.5}
\scriptsize
\caption{Effectiveness of Sentinel-1.}
\resizebox{0.6\linewidth}{!}{
\begin{tabular}{ccc}
\hline
 & + S1 & - S1 \\
\hline
mIoU (Seg) & 50.10 & 51.17 \\
\hline
\end{tabular}}
\label{tab:s1}
\end{wraptable}

\textbf{Limitations.} The FUSU dataset has three primary limitations. First, it is limited to five urban districts. Despite its rich geographical diversity and pixel data, including more global urban areas is desirable. We encourage the community to share high-quality, fine-grained land use datasets to advance urban monitoring. Second, land use change segmentation requires understanding human activities and production, unlike land cover, which directly corresponds to pixel values. Relying solely on remote sensing imagery makes high accuracy challenging. Third, as Table \ref{tab:s1} shows, because of the sensor gaps between optical images and SAR images, the simple concatenation of Sentinel-1 and Sentinel-2 is not ideal. Better fusion methods should be considered for synergizing both Sentinel-2 and Sentinel-1 strengths. In the future, we aim to design optical-SAR fusion methods and incorporate more multi-source data, such as economic and population data, to develop a multi-modal framework for comprehensive urban semantic understanding.

\textbf{Conclusion.} We present FUSU, a comprehensive multi-source, multi-temporal change segmentation dataset for fine-grained urban semantic understanding. FUSU includes a detailed 17-class land use classification system, 30 billion annotated pixels, 847 km² coverage, and temporal information from bi-temporal high-resolution images and monthly Sentinel data. This makes FUSU the most comprehensive urban semantic dataset available. We benchmark various methods to demonstrate FUSU's effectiveness in urban land use segmentation and change detection. Additionally, we introduce FUSU-Net, a model that fully utilizes the spatial, spectral, and temporal diversity of FUSU. We anticipate that FUSU and FUSU-Net will advance the development of powerful techniques for multi-source, multi-temporal change segmentation in urban environments without any negative societal impacts.

\bibliography{reference}
\newpage

\appendix

\section{Appendix}

\subsection{Annotations} \label{sec:ann}
The 17 land use classes used in FUSU are annotated according to the Chinese Land Use Classification Criteria (GB/T21010-2017) Level-1 classification system, i.e., \textit{traffic land, inland water, residential land, cropland, agriculture construction, blank, industrial land, orchard, park, public management, commercial land, public construction, special land, forest, storage, wetland, grass, background}. The detailed criteria and description of each class are shown in Table \ref{tab:criteria}. The annotation is conducted by two teams of geo-experts based on the ArcGIS geospatial software. Each team is responsible for one city and the annotation results are cross-checked by the other team. If there exists disagreement in some areas, these areas will be re-annotated when the agreement is reached. Leaders of two teams will randomly select 100 small areas in two cities for quality check. All objects are annotated as polygon features. Total annotation costs about 3 months. To ensure geographical continuity, the annotation is conducted on the full-scale images before image cropping.

\begin{table}[h]
\centering
\caption{Class description and criteria.}
\renewcommand{\arraystretch}{1.5}
\resizebox{\linewidth}{!}{
\begin{tabular}{ccl}
\hline
Value                & Name                                       & Criteria                                                                                                                                                                                                                                                                                                                                                                      \\ \hline
\textit{\textbf{1}}  & \textit{\textbf{Traffic land}}             & \begin{tabular}[c]{@{}l@{}}Refers to land for transportation facilities and their affiliated facilities such as railways, highways, airports, ports, \\ docks, pipelines, urban rail transit, various roads, and transport stations, excluding auxiliary roads and parking lots within other lands\end{tabular}                                                               \\ \hline
\textit{\textbf{2}}  & \textit{\textbf{Inland water}}             & \begin{tabular}[c]{@{}l@{}}Refers to natural land water bodies within the land area such as rivers, lakes, glaciers, and perennial snow, as well\\ as artificial land water bodies such as reservoirs, ponds, and canal water surfaces\end{tabular}                                                                                                                           \\ \hline
\textit{\textbf{3}}  & \textit{\textbf{Residential land}}         & Refers to urban and rural residential land and land for community service facilities supporting residential life                                                                                                                                                                                                                                                              \\ \hline
\textit{\textbf{4}}  & \textit{\textbf{Cropland}}                 & \begin{tabular}[c]{@{}l@{}}Refers to land mainly used for cultivating crops, with at least one crop cycle per year (including land used for \\ perennial crops cultivated in a manner of one or more crop cycles per year). This includes mature land, newly\\ developed, reclaimed, and organized land, fallow land (including fallow rotation and fallow land)\end{tabular} \\ \hline
\textit{\textbf{5}}  & \textit{\textbf{Agriculture construction}} & \begin{tabular}[c]{@{}l@{}}Refers to land where the surface cultivation layer has been destroyed for the service of agricultural production\\ and rural life, including rural roads and construction land for planting facilities, livestock and poultry facilities,\\ and aquaculture facilities\end{tabular}                                                                \\ \hline
\textit{\textbf{6}}  & \textit{\textbf{Blank}}                    & \begin{tabular}[c]{@{}l@{}}Refers to land within urban and village areas designated by national space planning with unclear planning use,\\ not to be developed within the planning period or to be developed under specific conditions\end{tabular}                                                                                                                          \\ \hline
\textit{\textbf{7}}  & \textit{\textbf{Industrial land}}          & Refers to land used for industrial and mining production                                                                                                                                                                                                                                                                                                                      \\ \hline
\textit{\textbf{8}}  & \textit{\textbf{Orchard}}                  & \begin{tabular}[c]{@{}l@{}}Refers to land used for cultivating perennial crops intensively for the collection of fruits, leaves, roots, stems, or\\ sap, with a coverage rate of more than 50\% or more than 70\% of the reasonable number of plants per acre, including land used for nurseries\end{tabular}                                                                 \\ \hline
\textit{\textbf{9}}  & \textit{\textbf{Park}}                     & \begin{tabular}[c]{@{}l@{}}Refers to land within urban and village construction areas for parks, protective green spaces, squares, and other\\ public open spaces, excluding auxiliary green spaces in other construction lands\end{tabular}                                                                                                                                  \\ \hline
\textit{\textbf{10}} & \textit{\textbf{Public management}}        & \begin{tabular}[c]{@{}l@{}}Refers to land for institutions and facilities of administrative bodies, groups, research, culture, education, sports,\\ health, social welfare, etc., excluding rural and urban community service facilities\end{tabular}                                                                                                                         \\ \hline
\textit{\textbf{11}} & \textit{\textbf{Commercial land}}          & Refers to land for commercial, business finance, and recreational facilities, excluding rural and urban community service facilities                                                                                                                                                                                                                                          \\ \hline
\textit{\textbf{12}} & \textit{\textbf{Public construction}}      & \begin{tabular}[c]{@{}l@{}}Refers to land for urban and regional infrastructure facilities such as water supply, drainage, power supply, gas\\ supply, heating, communication, postal services, broadcasting, sanitation, firefighting, main channels, and hydraulic works\end{tabular}                                                                                       \\ \hline
\textit{\textbf{13}} & \textit{\textbf{Special land}}                  & Refers to land for military, foreign affairs, religious, security, funeral purposes, and sites of historical relics with special properties                                                                                                                                                                                                                                   \\ \hline
\textit{\textbf{14}} & \textit{\textbf{Forest}}                   & \begin{tabular}[c]{@{}l@{}}Refers to land growing trees, bamboo, or shrubs. This does not include wetland growing trees, greening land\\ within urban and village areas, trees within the scope of railway and highway land, or trees for river and canal embankment protection\end{tabular}                                                                                  \\ \hline
\textit{\textbf{15}} & \textit{\textbf{Storage}}                  & Refers to land for logistics storage and strategic material reserve warehouses                                                                                                                                                                                                                                                                                                \\ \hline
\textit{\textbf{16}} & \textit{\textbf{Wetland}}                  & \begin{tabular}[c]{@{}l@{}}Refers to the land at the interface of land and water bodies where the water level is close to or at the surface,\\ or with shallow water layers, remaining in a natural state\end{tabular}                                                                                                                                                        \\ \hline
\textit{\textbf{17}} & \textit{\textbf{Grass}}                    & \begin{tabular}[c]{@{}l@{}}Refers to land mainly growing herbaceous plants, including sparse forest grasslands with a tree canopy density of less than 0.1\\ and shrub grasslands with shrub coverage of less than 40\%. This does not include wetlands or saline-alkali lands growing herbaceous plants\end{tabular}                                                         \\ \hline
\textit{\textbf{0}}  & \textit{\textbf{Background}}               & Others or extremely difficult to annotate                                                                                                                                                                                                                                                                                                                                     \\ \hline
\end{tabular}}
\label{tab:criteria}
\end{table}

\subsection{Sentinel Time Series}

\textbf{Sentinel-2.} The Sentinel-2 sensor is a multispectral sensor launched in 2015. The Sentinel-2 we use has 12 bands covering the VNIR and SWIR regions, with spatial resolutions of 10, 20, and 60 m. The swath width is 290 km. In general, the complete survey of the earth is repeated every 5 days. Here, we select all available Level-2A products (Bottom-Of-the-Atmosphere reflectances) in one single month, which are preprocessed through atmosphere correction, and compute the mean of these products to get the monthly-revisited observation data. All images are cloud-free by s2cloudless\footnote{https://github.com/sentinel-hub/sentinel2-cloud-detector}. Table \ref{tab:s2} summarizes the spectral and spatial attributes and applications of Sentinel-2 bands. Note that Sentinel-2 sensors have 10, 20, and 60 m spatial resolutions, and all bands are resampled to 10 m by the nearest interpolation method. 

\begin{table}[h]
\centering
\caption{Spectral and spatial attributes of Sentinel-2.}
\renewcommand{\arraystretch}{1.5}
\resizebox{\linewidth}{!}{
\begin{tabular}{ccccccc}
\hline
\begin{tabular}[c]{@{}c@{}}Original\\ band number\end{tabular} & \begin{tabular}[c]{@{}c@{}}FUSU\\ band number\end{tabular} & \begin{tabular}[c]{@{}c@{}}Band width\\ (mm)\end{tabular} & \begin{tabular}[c]{@{}c@{}}Center band\\ (mm)\end{tabular} & \begin{tabular}[c]{@{}c@{}}Original resolution\\ (m)\end{tabular} & \begin{tabular}[c]{@{}c@{}}FUSU resolution\\ (m)\end{tabular} & Usage                                         \\ \hline
1                                                              & 1                                                          & 20                                                        & 443                                                        & 60                                                                & 10                                                            & Atmospheric correction                        \\
2                                                              & 2                                                          & 65                                                        & 490                                                        & 10                                                                & 10                                                            & Vegetation aerosol scattering                 \\
3                                                              & 3                                                          & 35                                                        & 560                                                        & 10                                                                & 10                                                            & Green peak                                    \\
4                                                              & 4                                                          & 30                                                        & 665                                                        & 10                                                                & 10                                                            & Max chlorophyll absorption                    \\
5                                                              & 5                                                          & 15                                                        & 705                                                        & 20                                                                & 10                                                            & Not used in L2A context                       \\
6                                                              & 6                                                          & 15                                                        & 740                                                        & 20                                                                & 10                                                            & Not used in L2A context                       \\
7                                                              & 7                                                          & 20                                                        & 783                                                        & 20                                                                & 10                                                            & Not used in L2A context                       \\
8                                                              & 8                                                          & 115                                                       & 842                                                        & 10                                                                & 10                                                            & {\color[HTML]{1F1F1F} LAI}                    \\
8a                                                             & 9                                                          & 20                                                        & 865                                                        & 20                                                                & 10                                                            & Water vapor absorption reference              \\
9                                                              & 10                                                         & 20                                                        & 945                                                        & 60                                                                & 10                                                            & Water vapor absorption atmospheric correction \\
11                                                             & 11                                                         & 90                                                        & 1610                                                       & 20                                                                & 10                                                            & Soils detection                               \\
12                                                             & 12                                                         & 180                                                       & 2190                                                       & 20                                                                & 10                                                            & AOT determination                             \\ \hline
\end{tabular}}
\label{tab:s2}
\end{table}

\textbf{Sentinel-1.}The Sentinel-1 mission provides data from a dual-polarization C-band Synthetic Aperture Radar (SAR) instrument at 5.405GHz (C band). The satellites are to operate day-and-night and perform a synthetic aperture with radar imaging in all weather conditions. Sentinel-1 images in FUSU have 2 bands VV and VH (dual-band cross-polarization, vertical transmit/horizontal receive). The revisited cycle is 6 days. To get the monthly-revisited data, we also first select all available products in one single month and process the raw data by noise removal, radiometric calibration, and orthorectification, and then compute the mean of these products. 
\subsection{Dataset Expanding and Benefits} \label{sec:expand}
\begin{figure}[h]
    \centering
    \includegraphics[width=0.95\linewidth]{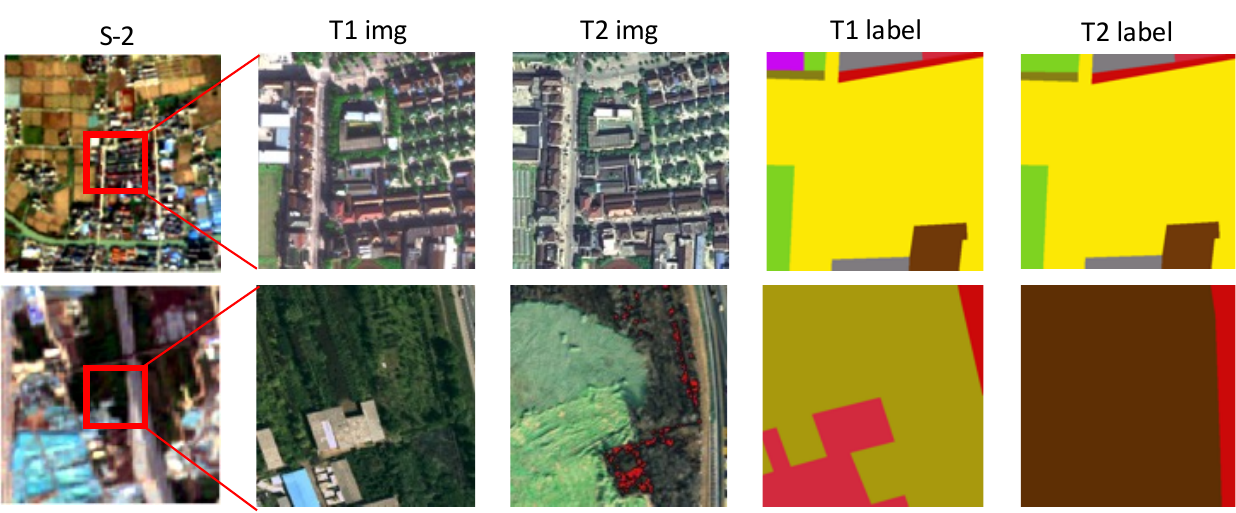}
    \caption{The alignment of Google Earth images and Sentinel images.}
    \label{fig:align}
\end{figure}

\textbf{Dataset expanding.} To expand the temporal information of FUSU, we develop a data-expanding paradigm that combines the temporally rich Sentinel images with high-resolution Google Earth images. This method involves three steps. First, we crop Google Earth images into $512\times512$ patches and generate a shapefile that is five times larger centered around the patch's midpoint for each patch. Then these shapefiles are used to download Sentinel images from Google Earth Engine. The time series of Sentinel images span the entire time interval between the snapshot times of two Google Earth images, with one image per month. Sentinel-1 images are preprocessed by noise removal, calibration, and correction. We then process Sentinel-2 to usable conditions by cloud removal, atmosphere correction, radiometric calibration, and orthorectification.
As a result, Sentinel images have a size of $128\times128$ with a resolution of 10 m. Note that the Sentinel patches and Google Earth patches are not strictly aligned. The geographic content covered by a Google Earth patch only occupies the central areas of the corresponding Sentinel patch as shown in Fig. \ref{fig:align}. This consideration is adopted for two main reasons: First, if strict alignment were enforced, the Sentinel patch size would be very small due to the significant resolution difference, resulting in insufficient semantic information for model training. Second, the larger coverage area of Sentinel patches captures the surrounding context and landscape variations and helps identify and understand patterns and trends in broader spatial dynamics. 

This data-expanding paradigm enhances FUSU's temporal resolution, capturing more detailed changes during time series. Moreover, it is versatile enough to be extended to other readily available change detection datasets. We will provide the process steps and code accordingly for the community.

\textbf{Benefits.} Supplementing bi-temporal high-resolution images with the public multi-temporal Sentinel-2 and Sentinel-1 images has benefits in both clear geographic feature awareness and feature change awareness. First, Sentinel images provide high temporal-resolution observations, filling the gap of continuous temporal information between the snapshot times of bi-temporal images. This enables the capture of monthly changes and enhances the ability to detect and understand changes over time. Second, Sentinel images offer extensive spatial information. Thanks to the data-expansion design described in Sec. \ref{sec:expand}, each Sentinel image is centered around the corresponding high-resolution image. This additional spatial information provides larger receptive fields for our model, ensuring geographical continuity and enabling broader area observations. Third, Sentinel-1 and Sentinel-2 images provide diverse observations from different modalities. These varied modalities enhance the dataset by capturing a wider range of features and details, providing multiple ways of observations on different kinds of human activities.

\subsection{License}
Use of the Google Earth images must respect the "Google Earth" terms of use. All images and their associated annotations in FUSU can be used for academic purposes only, and any commercial use is prohibited (CC BY-NC-SA 4.0).

\subsection{Extra Experiment Results} \label{sec:exp}
\subsubsection{Dataset Split}
We show our training and testing dataset split in the URL link. In general, table \ref{tab:split} shows the details. Note that for change detection, we only select the patches that have changed pixels.

\begin{table}[]
\centering
\caption{Training and testing data split.}
\renewcommand{\arraystretch}{1.5}
\resizebox{0.5\linewidth}{!}{
\begin{tabular}{cccc}
\hline
Data             & Train  & Test   & Val   \\ \hline
Complete         & 43,927 & 12,550 & 6,275 \\
Xi'an            & 25,303 & 7,205  & 3,660 \\
Jiaxing          & 18,624 & 5,345  & 2,615 \\
Change Detection & 16,998 & 4,813  & 2,413 \\ \hline
\end{tabular}}
\label{tab:split}
\end{table}

\begin{figure}[h]
    \centering
    \includegraphics[width=0.4\linewidth]{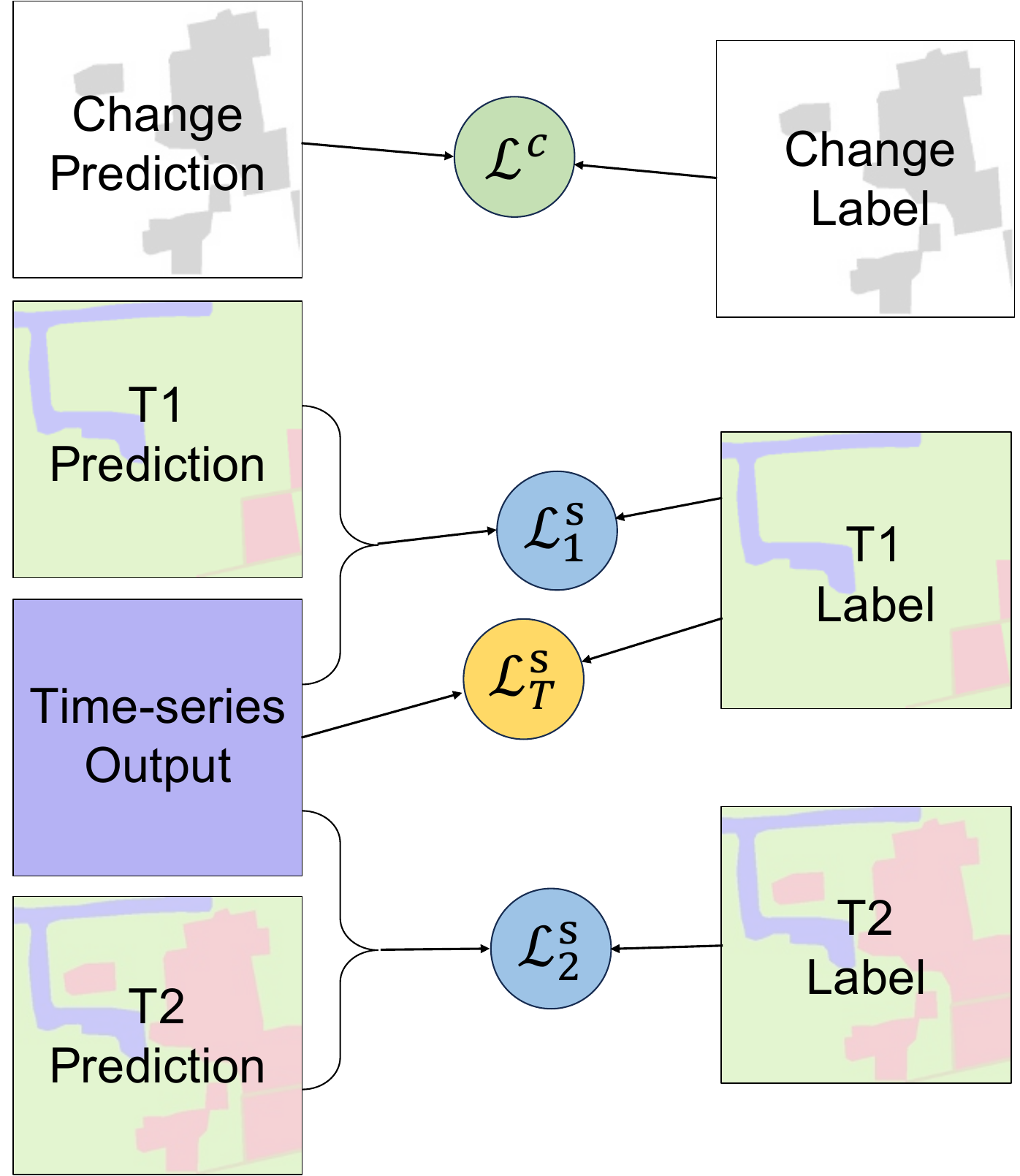}
    \caption{The visualization of supervision in FUSU-Net.}
    \label{fig:supervision}
\end{figure}

\subsubsection{Implementation Details} \label{sec:detail}
For segmentation, we use a Stochastic Gradient Descent (SGD) optimizer with a momentum of 0.9 and a weight decay of 1e-4. The learning rate is 0.01, and a 'poly' scheduler with power 0.9 is applied. The batch size is 8 and the max training iterations are 80$k$. For semantic change detection, we use AdamW as our optimizer and $\beta_1$ is 0.5 and $\beta_2$ is 0.999. The learning rate is 3e-4 and linearly decays are applied to 0 until trained for max epochs. The batch size is 8 and the max training epochs are 200. For binary change detection, the learning rate is 0.001, and other settings are the same as the semantic change detection. For FUSU-Net, the settings are the same as other methods in different tasks. For domain adaptation, we adopt the architectures in \cite{wang2021loveda} and keep the default settings. All experiments are conducted on 4 NVIDIA GTX 4090 GPUs with 25 GB memory.

\subsubsection{FUSU-Net Supervision} \label{sec:supervision}

Fig. \ref{fig:supervision} shows the supervision in FUSU-Net. There are four outputs of FUSU-Net, i.e., change prediction, T1 prediction, time-series prediction, T2 prediction, and three labels, i.e., change label, T1 label and T2 label. To balance segmentation and change detection, we set the weight of change loss as 2, and the weights of segmentation losses as 1. The total loss function is calculated as:
\begin{equation}
    \mathcal{L} = \mathcal{L}_{1}^s + \mathcal{L}_{2}^s + \mathcal{L}_{T}^s + 2 \times \mathcal{L}^c  \label{4}
\end{equation}

\subsubsection{Domain Adaptation}
Because of the feature gaps between Jiaxing and Xi'an, as we discussed in Sec. \ref{sec:fusu} and Sec. \ref{sec:seg}, the FUSU dataset also has the ability to support domain adaptation. Here we evaluate the performance of 6 unsupervised domain adaptation methods on FUSU dataset, which include FADA \cite{wang2020classes}, PyCDA \cite{lian2019constructing}, CLAN \cite{luo2019taking}, CBST \cite{zou2018unsupervised}, AdaptSeg \cite{tsai2018learning} and IAST \cite{mei2020instance}. IoU per class and mIoU over 17 classes are calculated.

\begin{table}[]
\centering
\caption{Domain adaptation results obtained from training and testing on the whole dataset.}
\renewcommand{\arraystretch}{1.5}
\resizebox{\linewidth}{!}{
\begin{tabular}{cccccccccccccccccccc}
\hline
\multirow{2}{*}{Domain}                                                  & \multirow{2}{*}{Method} & \multicolumn{17}{c}{IoU per class (\%)}                                   & \multirow{2}{*}{mIoU} \\ \cline{3-19}
                                                                         &                         & 1 & 2 & 3 & 4 & 5 & 6 & 7 & 8 & 9 & 10 & 11 & 12 & 13 & 14 & 15 & 16 & 17 &                       \\ \hline
\multirow{6}{*}{\begin{tabular}[c]{@{}c@{}}Jiaxing\\ $\downarrow$ \\ Xi'an\end{tabular}} 
& FADA \cite{wang2020classes}                 & 5.23 & 33.17 & 6.32 & 5.10 & 11.45 & 32.67 & 3.14 & 6.28 & 9.12 & 3.24 & 1.02 & 1.08 & 5.18 & 8.33 & 0.20 & 0.11 & 1.15 & 8.23                    \\
&PyCDA \cite{lian2019constructing}               & 3.08 & 32.54 & 8.11 & 1.28 & 12.35 & 26.48 & 8.27 & 6.17 & 10.33 & 5.42 & 1.10 & 1.22 & 9.18 & 3.21 & 1.12 & 1.08 & 1.16 & 8.34                     \\
& CLAN \cite{luo2019taking}                 & 4.33 & 24.67 & 6.42 & 3.15 & 18.34 & 39.78 & 4.22 & 8.17 & 14.23 & 4.14 & 2.08 & 3.18 & 3.11 & 6.33 & 1.16 & 1.41 & 1.57 & 9.63                      \\
& CBST \cite{zou2018unsupervised}        & 17.21 & 10.43 & 5.34 & 2.11 & 28.78 & 39.32 & 2.07 & 5.38 & 7.22 & 3.42 & 7.12 & 2.03 & 2.14 & 2.21 & 0.44 & 0.45 & 1.13 & 9.24                      \\
& AdaptSeg  \cite{tsai2018learning}   & 7.23   & 35.78  &  14.54 &   3.12 & 13.67 & 25.22  & 3.18  & 6.34  & 8.28  & 5.13   & 2.11   & 3.12   & 3.14   & 2.25   & 1.15   & 1.19   &  1.24  &  9.68                     \\
& IAST   \cite{mei2020instance}                 & 8.13   & 36.42  & 15.78 & 4.12  & 15.89  & 27.65  & 5.12  & 7.42  & 10.21  & 6.38  & 3.21  & 4.19  & 4.33  & 8.45  & 2.81  & 2.22  & 2.35  & 10.93                      \\ \hline
\multirow{6}{*}{\begin{tabular}[c]{@{}c@{}}Xi'an\\ $\downarrow$ \\Jiaxing\end{tabular}} 
& FADA  \cite{wang2020classes}                  & 11.38 & 31.67 & 10.21 & 4.18 & 11.24 & 22.53 & 9.12 & 10.28 & 7.18 & 2.14 & 1.11 & 1.22 & 3.23 & 1.08 & 1.14 & 1.49  & 1.25        & 9.97              \\
& PyCDA \cite{lian2019constructing}                  & 13.45 & 10.24 & 7.34 & 2.13 & 3.28 & 25.65 & 3.14 & 2.15 & 9.38 & 2.08 & 1.12 & 1.18 & 5.23 & 3.15 & 1.77 & 2.02 & 1.14 & 8.70                         \\
& CLAN \cite{luo2019taking}                   & 11.38 & 13.42 & 12.24 & 4.28 & 10.53 & 17.68 & 4.17 & 8.28 & 13.42 & 2.14 & 1.11 & 1.25 & 5.21 & 1.08 & 0.78 & 1.34 & 1.21 & 9.24                       \\
& CBST  \cite{zou2018unsupervised}                  & 6.42 & 23.78 & 20.12 & 2.11 & 28.89 & 22.34 & 3.21 & 14.27 & 17.45 & 2.14 & 1.23 & 1.28 & 3.14 & 1.16 & 0.62 & 1.39 & 1.18 & 10.21                     \\
& AdaptSeg \cite{tsai2018learning}               & 6.28  & 40.18   & 6.11  & 2.17  & 18.42  & 40.78  & 2.13   & 6.42  & 4.11  & 2.18   & 1.21   & 1.27   & 1.34   & 1.16   & 1.24   & 1.29   & 1.35   & 10.63                      \\
& IAST     \cite{mei2020instance}                 & 8.38   & 38.27  & 8.12  & 3.24  & 16.42  & 38.68  & 3.24  & 7.24  & 4.28  & 3.15  & 1.18  & 1.25  & 2.27  & 2.17  & 2.23  & 2.28  & 2.32  & 10.89                      \\ \hline
\end{tabular}}
\label{tab:da}
\end{table}

\begin{table}[]
\centering
\caption{Top performances compared with other datasets.}
\renewcommand{\arraystretch}{1.5}
\resizebox{0.2\linewidth}{!}{
\begin{tabular}{cc}
\hline
Dataset             & mIoU   \\ \hline
GID \cite{GID2020} & 90.79 \\
ISPRS Potsdam\footnote{https://www.isprs.org/education/benchmarks/UrbanSemLab/2d-sem-label-potsdam.aspx} & 82.17 \\
LoveDA \cite{wang2021loveda} & 49.02 \\
FLAIR \cite{garioud2024flair} & 54.51 \\
FUSU & 46.88 \\
\hline
\end{tabular}}
\label{tab:comp}
\end{table} 

\textbf{Overall results.} Table \ref{tab:da} shows the semantic change detection results. We can see there are some
easy classes for all unsupervised domain adaptation methods (i.e., inland water-2, blank-6), which are similar to the results of semantic segmentation. Some classes bring challenges (i.e., storage-15, grass-17), indicating difficulty in adapting to feature changes in those specific categories. Also, we can see the interchange between the source domain and the target domain will affect the performance of domain adaptation tasks. Xi'an to Jiaxing task gets higher performance on blank-5 than the Jiaxing to Xi'an task. There isn't much disparity in performance
 between two mainstream approaches, i.e., adversarial
training (AdaptSeg, CLAN, FADA) and self-training (CBST, PyCDA). In summary, these methods get unsatisfactory performance on our dataset. The results show little improvement compared to the source-only results, and in some cases, they are even worse. We hypothesize the following two reasons. First, general domain adaptation methods in the field of computer vision cannot adapt to the domain characteristics of the FUSU dataset, necessitating the development of improved methods. A customized method might achieve better results. Second, the categories in Jiaxing and Xi'an are discontinuous, with Jiaxing having more cropland and Xi'an having more urban buildings, resulting in a significant domain gap. This large gap makes it challenging for the methods to learn effectively.

\subsubsection{Comparison with Other Datasets}
We investigate the difficulty of several open-source segmentation datasets by comparing the top performances on these datasets via HRNet. We can see from Table \ref{tab:comp} that FUSU has the lowest performance among all datasets, which indicates the difficulty of the FUSU dataset. We summarize the challenges of FUSU from two aspects. First, the feature gaps between Jiaxing and Xi'an increase the difficulty of this dataset. The training model must adapt with two main features during one end-to-end period. Second, the land use classification involves many understandings of human activities and production rather than land cover, which can directly correspond to pixel values. Therefore, relying solely on remote sensing imagery makes achieving high accuracy challenging. Multi-source data is needed for better performance.

\begin{figure}[h]
    \centering
    \includegraphics[width=0.8\linewidth]{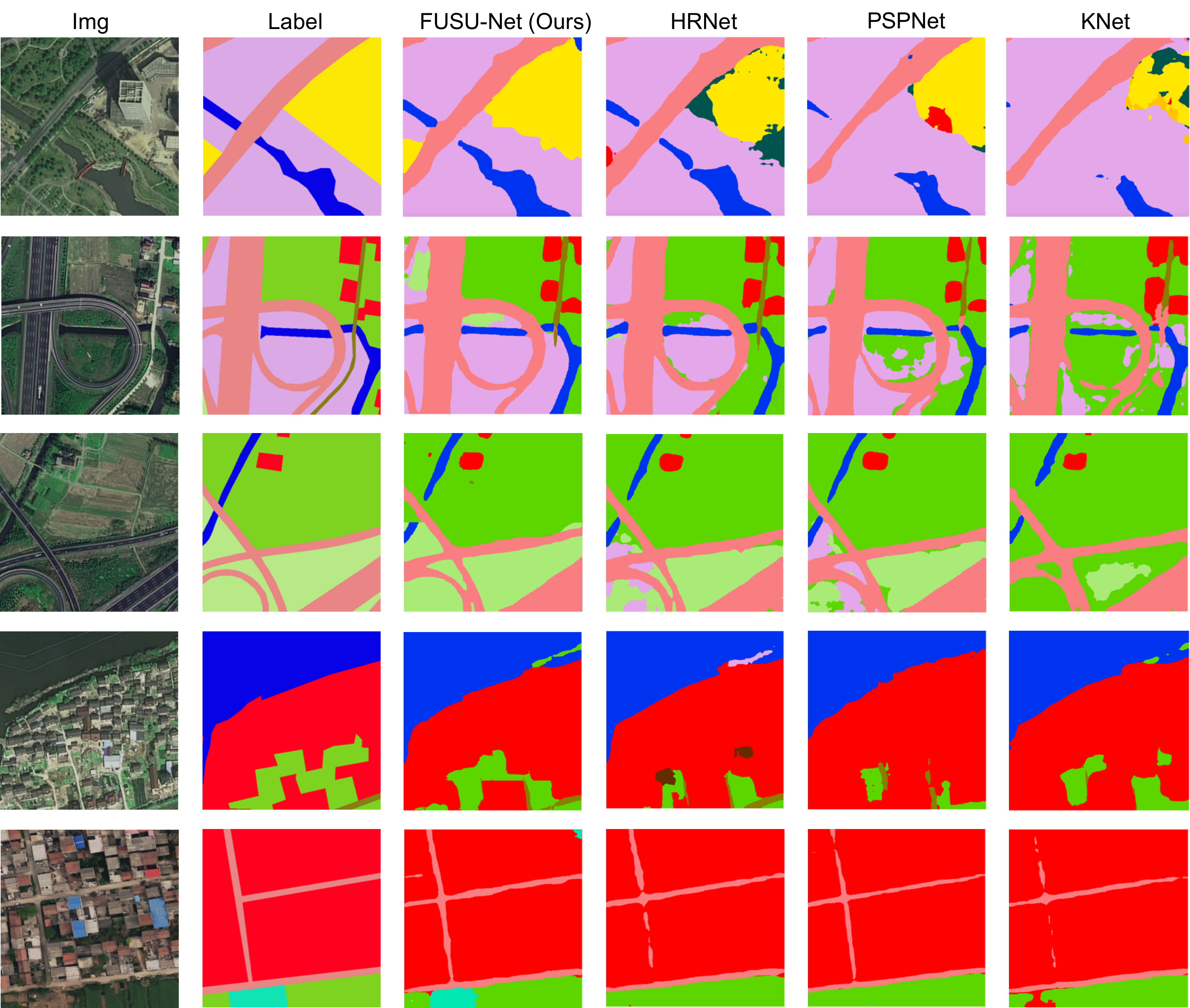}
    \caption{The visualization of results.}
    \label{fig:visual}
\end{figure}
\end{document}